\documentclass[letterpaper, 10 pt, conference]{ieeeconf}  

\IEEEoverridecommandlockouts                              

\usepackage{cite}
\usepackage{graphicx}
\usepackage[font=small,labelfont=bf]{caption}
\usepackage[font=small,labelfont=bf]{subcaption}
\usepackage{float}
\usepackage[T1]{fontenc}
\usepackage{multirow}
\usepackage{enumerate}
\usepackage{amsmath}
\usepackage{adjustbox}
\usepackage{threeparttable}
\usepackage{wrapfig}
\usepackage{xspace}
\usepackage{xcolor}   
\usepackage{pifont}    
\definecolor{formalgreen}{rgb}{0.1, 0.7, 0.1}  
\definecolor{formalred}{rgb}{0.9, 0.2, 0.2}  
\newcommand{\cmark}{\textcolor{formalgreen}{\checkmark}}  
\newcommand{\xmark}{\textcolor{formalred}{\ding{55}}}           

\newcommand{\acronyme}{GOAG\xspace}
\newcommand{\acronymefull}{Generative and Object-Agnostic Grasp Planner for Dexterous Robotic Manipulation\xspace}

\usepackage[dvipsnames]{xcolor}
\newcommand{\red}[1]{{\color{red}#1}}
\newcommand{\blue}[1]{{\color{blue}#1}}

\definecolor{lightblue}{RGB}{100, 149, 237}

\usepackage{colortbl}
\usepackage{multirow}
\usepackage[para]{footmisc}
\usepackage{lipsum}
\usepackage{booktabs}

\usepackage{xspace}
\usepackage{hyperref}

\makeatletter
\DeclareRobustCommand\onedot{\futurelet\@let@token\@onedot}
\def\@onedot{\ifx\@let@token.\else.\null\fi\xspace}

\makeatother

\usepackage{amsmath}
\usepackage{amssymb}
\usepackage{dsfont}
\usepackage{etoolbox}
\usepackage{color}
\usepackage{ulem}

\newif\ifshowedits

\newcommand{\addeditor}[3]{%
  \definecolor{#1color}{rgb}{#3}
  \expandafter\newcommand\csname #1\endcsname[1]{%
  \ifshowedits
    {\color{#1color} ##1}%
  \else
    {##1}%
  \fi
  }%
  \expandafter\newcommand\csname #1rmk\endcsname[1]{%
  \ifshowedits
    {\color{#1color} {\bf [#2: ##1]}}
  \fi
  }%
  \expandafter\newcommand\csname #1rpl\endcsname[2]{%
  \ifshowedits
    {\color{#1color} ##1 \sout{##2}}
  \else
    {##1}
  \fi
  }%
}

\newcommand{\createtextvar}[1]{
  \expandafter\newcommand\csname #1\endcsname{%
  {\text{#1}}
}%
}
\newcommand{\mycomment}[1]{}

\newcommand{\calC}{{\cal C}}
\newcommand{\calD}{{\cal D}}

\newcommand{\calF}{{\cal F}}

\newcommand{\calH}{{\cal H}}

\newcommand{\calN}{{\cal N}}
\newcommand{\calO}{{\cal O}}

\newcommand{\vcomment}[1]{}

\addeditor{julien}{JM}{1.0, 0.349, 0.0}
\addeditor{boris}{BM}{0.0, 0.5, 0.0}
\addeditor{mathieu}{MG}{0,0,1}
\addeditor{liming}{LC}{1.0, 0.5, 1.0}
\addeditor{template}{TP}{1,0,0}
\showeditstrue

\title{\LARGE \bf \acronyme: \acronymefull}

\author{Julien Mérand$^{1}$, Boris Meden$^{1}$, Mathieu Grossard$^{1}$ and Liming Chen$^{2}$%
\thanks{$^{1}$Université Paris-Saclay, CEA, List, F-91120 Palaiseau, France}%
\thanks{$^{2}$Ecole Centrale de Lyon, CNRS, LIRIS, UMR5205, Institut Universitaire de France (IUF), F-69130 Ecully, France}%
\thanks{This work has been partially supported by TIRREX ANR-21-ESRE-0015. Experiments presented in this paper were carried out thanks to a platform of the Robotex 2.0 French research infrastructure.
The authors would like to thank Fabien Spindler and Romain Lagneau for their valuable assistance in achieving the experimental results presented in this paper.
Liming Chen in this research  was in part supported by the French Research Agency ANR, l’Agence Nationale de Recherche, through the projects Aristotle (ANR-21-FAI1-0009-01),  Astérix (ANR-23-EDIA-0002), DEMETER  (ANR-25-HTCE-0002) and PROTEUS (ANR-25-TSIA-0011-01), the French national investment prioritary program through the PSPC FAIR WASTE project.
This project has received funding from the European Union’s Horizon Europe research and innovation program under grant agreement nº 101135708 (JARVIS Project).}
}

\begin{document}

\maketitle
\thispagestyle{empty}
\pagestyle{empty}

\begin{abstract}

Multifingered grasping is a crucial robotic skill, but current deep-learning grasp planners often struggle to generalize to new objects because they are trained on limited, object-specific datasets. We introduce a fundamentally different approach, grounded in the observation that the gripper and the object share identical surface geometry at their mutual contact points. We propose \acronyme: \acronymefull, a novel deep generative model that learns a compact latent representation of a specific gripper's contact surface distribution, enabling the efficient sampling of valid grasp configurations without relying on object-specific training data. We show that by introducing object features only at inference time, our model can effectively retrieve admissible contact areas that are compatible with the gripper’s capabilities. We validate our approach through extensive experiments on established grasp protocols in both simulated and real-world scenarios, demonstrating its effectiveness with different grippers from the literature.
Our method delivers state-of-the-art results on the objects from the MultiDex dataset, achieving an average success rate of $86.93\%$. It offers significantly faster processing when generating numerous grasps, while matching the performance of leading approaches specifically trained on this dataset. Unlike these methods, our approach does not rely on 
object-specific training data, highlighting the advantages of object-agnostic learning. It effectively addresses the generalization challenges faced by traditional data-driven grasp planners. Code and videos are available on our project website \blue{\url{https://cea-list.github.io/goagweb/}}.

\end{abstract}

\section{Introduction}

Dexterous grasping remains a highly complex and largely unsolved challenge for multi-fingered robotic hands. While sophisticated hardware like Allegro~\cite{allegro}, ShadowHand~\cite{tuffield2003shadow}, and Barrett~\cite{townsend2000barretthand} exists, fully unlocking their potential requires grasp planners that can match their kinematic complexity in real-time. Traditional analytical methods like GraspIt!~\cite{miller2004graspit} struggle with these time constraints. Conversely, modern data-driven approaches offer faster inference but typically evaluate performance on restricted sets of objects, severely limiting their generalization to novel, real-world shapes.

\begin{figure}[!t]
    \centering
    \includegraphics[width=\linewidth]{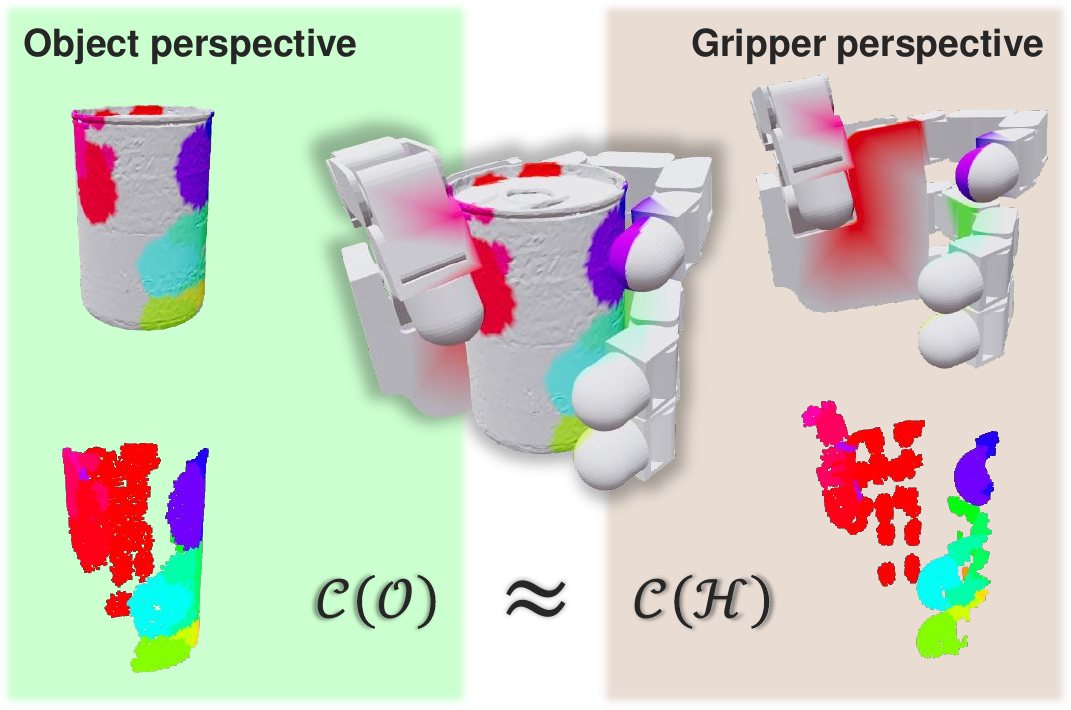}
    \caption{\textbf{\acronyme Paradigm.} A successful grasp on an object induces dual contact zones on both object and gripper, at the intersection of the two geometries. Our method is built on this key observation: these contact zones ($\calC(.)$) are closely the same from either perspective. \acronyme capitalizes on this by training exclusively on gripper geometry, allowing it to learn a robust and generalizable grasping strategy without ever being exposed to a grasp database with specific objects geometries.}
    \label{fig:intro_teaser}
\end{figure}

Our research addresses this generalization bottleneck by shifting from the common grasp-centric or object-centric paradigms to a gripper-centric perspective. As illustrated in Figure~\ref{fig:intro_teaser}, we aim to capture the combinatorial possibilities of contact areas inherent to a specific gripper design prior to seeing any object. To this end, we formulate the first data-driven grasp planner that is object-agnostic at training time, meaning no object geometry is utilized during training, named \acronyme (\acronymefull).

\begin{table*}[thbp]
    \resizebox{\textwidth}{!}{
        \begin{tabular}{@{}lcccccccc@{}}
            \toprule
            & \begin{tabular}[c]{@{}c@{}}Grasp\\ Representation\end{tabular} & \begin{tabular}[c]{@{}c@{}}Gripper\\ Pose\end{tabular} & \begin{tabular}[c]{@{}c@{}}Gripper\\ Joint Values\end{tabular} & \begin{tabular}[c]{@{}c@{}}Force\\ Closure\end{tabular} & \begin{tabular}[c]{@{}c@{}}Non-\\ Penetration\end{tabular} & \begin{tabular}[c]{@{}c@{}}Training\\ Set\end{tabular} & \begin{tabular}[c]{@{}c@{}}Working\\ Reference Frame\end{tabular} & \begin{tabular}[c]{@{}c@{}}Optional Grasp\\ Preference Interface\end{tabular} \\
            \midrule
            DFC \cite{liu2021synthesizing} & Direct & Optimized & Optimized & \cmark & \cmark & No & Object & \xmark \\
            UniGrasp \cite{shao2020unigrasp} & Intermediate & IK Solved & IK Solved & \cmark & \xmark & Objects + Grippers & Object & \xmark \\
            GeoMatch \cite{attarian2023geometry} & Intermediate & IK Solved & IK Solved & \cmark  & \xmark & Objects + Grippers & Object & \xmark \\
            GenDexGrasp \cite{li2023gendexgrasp} & Intermediate & Optimized & Optimized & \cmark & \cmark & Objects + Grippers & Object & \xmark \\
            ManiFM~\cite{xu2024manifoundation} & Intermediate & Optimized & Optimized & \cmark & \cmark & Objects + Grippers & Object & Contact Region \\
            DRO-Grasp~\cite{wei2024d} & Intermediate & Optimized & Optimized & \cmark & \cmark & Objects + Grippers & Object & Palm Orientation \\ 
            DexDiffuser~\cite{weng2024dexdiffuser} & Direct & Learned & Learned & A posteriori & A posteriori & Objects + Grippers & Object & \xmark \\
            DexGrasp Anything~\cite{zhong2025dexgrasp} & Direct & Learned & Learned & \cmark & \cmark & Objects + Grippers & Object & \xmark \\
            \textbf{\acronyme (Ours)} & Intermediate & Sampled & Optimized & \cmark & \cmark & Gripper Only & Gripper & Palm Full Pose \\ 
            \bottomrule
        \end{tabular}
    }
    \caption{\textbf{Comparison of Dexterous Grasp Planning Methods.}}
    \label{tab:relatedgrasp}
    \vspace{-15pt}
\end{table*}

Instead of learning from a set of stable grasps achieved on a finite set of object geometries, \acronyme learns the intrinsic distribution of feasible contact areas from a synthetic dataset that maps grasp taxonomy labels~\cite{escorcia2023task} (Figure \ref{fig:grasp_types}) to randomly sampled kinematic configurations. 

Specifically, we train a Conditional Variational Auto Encoder (CVAE)~\cite{sohn2015learning} to learn the conditional distribution of these contact points. This strategy makes the model's training inherently object-agnostic, as it relies exclusively on the gripper’s intrinsic geometry and kinematics, enabling zero-shot generalization to arbitrary object shapes by introducing object features solely at inference time. The model's input is a Basis Point Set (BPS)~\cite{prokudin2019efficient} encoding of a point cloud. During inference, using the same BPS encoder, the model adeptly retrieves potential contact areas on the object shape that are compatible with the gripper’s kinematics. An additional PointNet++~\cite{qi2017pointnet++} network then associates these identified contact zones with specific gripper links, and the gripper joints are subsequently optimized to solve this mapping problem, enabling precise grasp execution.
In summary, our contributions are the following:
\textbf{1) Object-Agnostic Learning Strategy.} We introduce a novel training paradigm for dexterous grasping that relies exclusively on the gripper’s intrinsic geometry and kinematics. By decoupling the learning phase from object data, we eliminate the bias toward specific object datasets inherent in traditional methods.
\textbf{2) Demonstrated Efficiency and Generalization.} We provide extensive experimental validation in both simulated and real-world environments. By following established evaluation protocols~\cite{wei2024d} and extending testing to novel objects, we demonstrate our approach's generalization compared to existing methods.

\section{Related Work}

Recent advancements in dexterous grasp planning have been largely propelled by data-driven approaches~\cite{newbury2023deep}. Table~\ref{tab:relatedgrasp} summarizes their characteristics.

\subsection{Grasp Databases}

The success of modern data-driven models heavily relies on large-scale grasp databases. While early datasets relied on time-intensive human demonstrations~\cite{eiband2021identification}, motion capture~\cite{brahmbhatt2020contactpose, taheri2020grab}, teleoperation~\cite{liu2024realdex} or even thermal imaging~\cite{brahmbhatt2019contactdb, brahmbhatt2020contactpose}, recent works have pioneered the automated, synthetic generation of thousands of physically validated candidate grasps~\cite{wang2023dexgraspnet, turpin2023fast, xu2023unidexgrasp, li2023gendexgrasp, zhong2025dexgrasp}, based on analytical grasp planners~\cite{liu2021synthesizing, miller2004graspit} and validating stability using physically realistic simulators like Isaac Gym~\cite{makoviychuk2021isaac}.\\
These databases are costly to generate and because they map specific grippers to specific objects, models trained on them remain intrinsically biased by the training geometry, hindering true generalization to unseen shapes.

\subsection{Learning Explicit Grasps}

A primary family of approaches directly learns the mapping from an object's representation to a complete grasp configuration, including the gripper's pose and joint values. This category spans from early regression models~\cite{liu2020deep, xu2024dexterous} to sophisticated generative architectures by, modeling the conditional probability distribution~\cite{jiang2021hand, mayer2022ffhnet}, using diffusion models~\cite{weng2024dexdiffuser, huang2023diffusion, lu2024ugg}, or foundation models~\cite{xu2024manifoundation, zhong2025dexgrasp, wei2024grasp}. \\
Despite their expressiveness and ability to process complex point clouds, these direct-prediction methods share a common limitation: the generated poses do not always inherently satisfy physical constraints. They frequently require a computationally expensive post-hoc validation step or a learned discriminator to filter out penetrating or unstable grasps.

\subsection{Learning Intermediate Representation}

Instead of directly predicting a grasp pose, alternative methods learn an intermediate representation, most commonly a contact map defining desired grasp regions on the object's surface~\cite{attarian2023geometry, jiang2021hand, li2023gendexgrasp, khargonkar2025robotfingerprint}. The final grasp pose is then derived using an inverse kinematics solver, which ensures non-penetration and inherits stability from the training data.
Other methods extract gripper-specific features to to tackle the multi-embodiment challenge~\cite{wei2024d, shao2020unigrasp, xu2021adagrasp}. \\
While these intermediate approaches improve physical validity, they still depend on pre-existing object-grasp databases for training. As highlighted in Table~\ref{tab:relatedgrasp}, \acronyme fundamentally breaks from this paradigm. By modeling the gripper's contact capabilities entirely independently of object data, we achieve an object-agnostic training process that does not require pre-computed grasp examples, which are often costly to generate.

\section{Method}
\label{sec:method}

\begin{figure}[t]
    \centering
    \includegraphics[width=0.7\columnwidth]{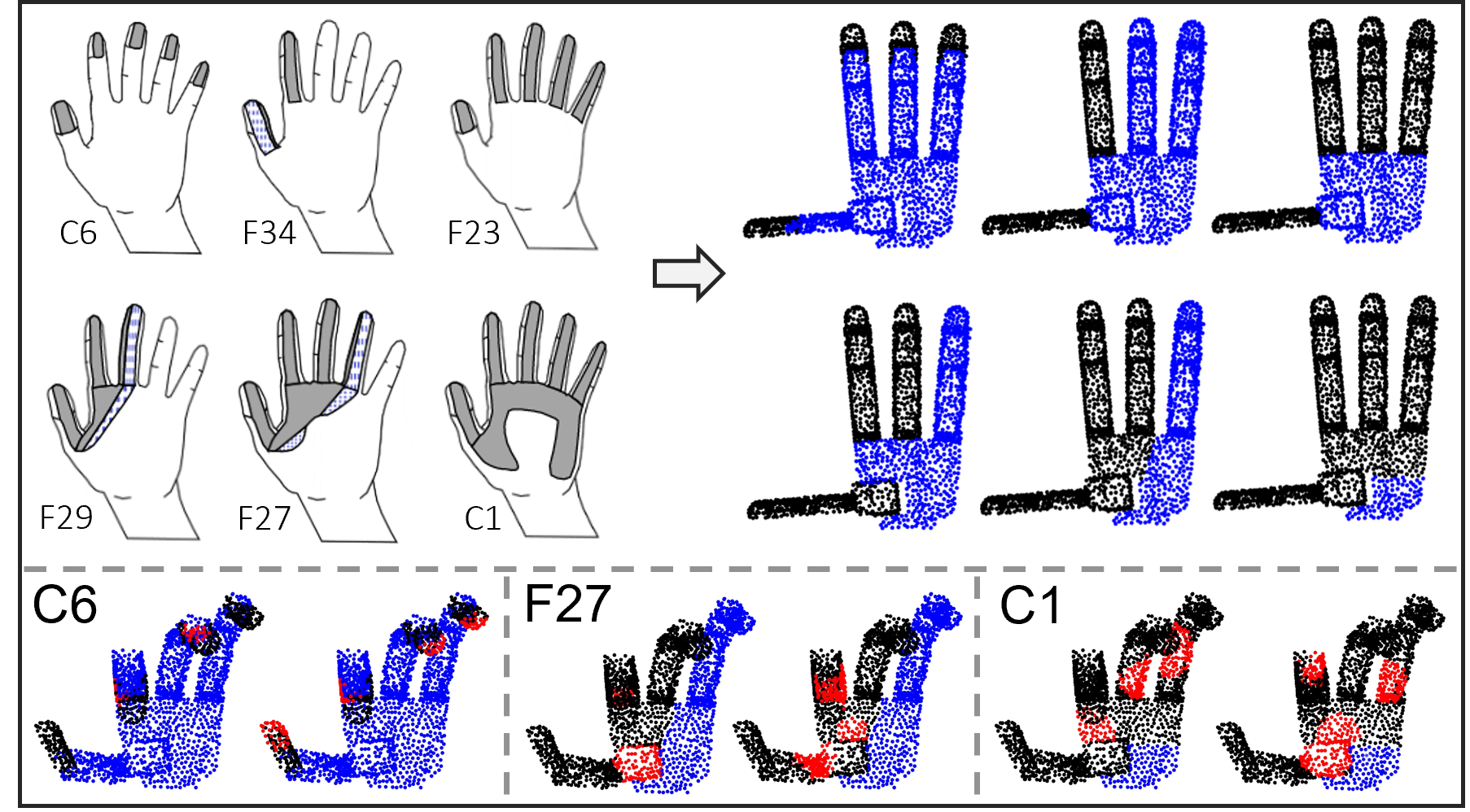}
    \caption{\textbf{Grasp Taxonomy Adaptation and Contact Sampling. (Top)} We adapt the human grasp taxonomy from \cite{escorcia2023task} to the Allegro Hand geometry. For each grasp type (e.g., C6, F27), we define a corresponding \textbf{admissible contact region} (black points), distinguishing it from the \blue{non-contact surface} (blue points). \textbf{(Bottom)} Data generation mechanism: We randomly sample specific \red{contact points} (red) strictly within the admissible black regions. This allows the model to learn structured, feasible contact distributions based solely on gripper kinematics, independent of any object.}
    \label{fig:grasp_types}
    \vspace{-15pt}
\end{figure}

\begin{figure*}[!ht]
    \centering
    \includegraphics[width=\textwidth]{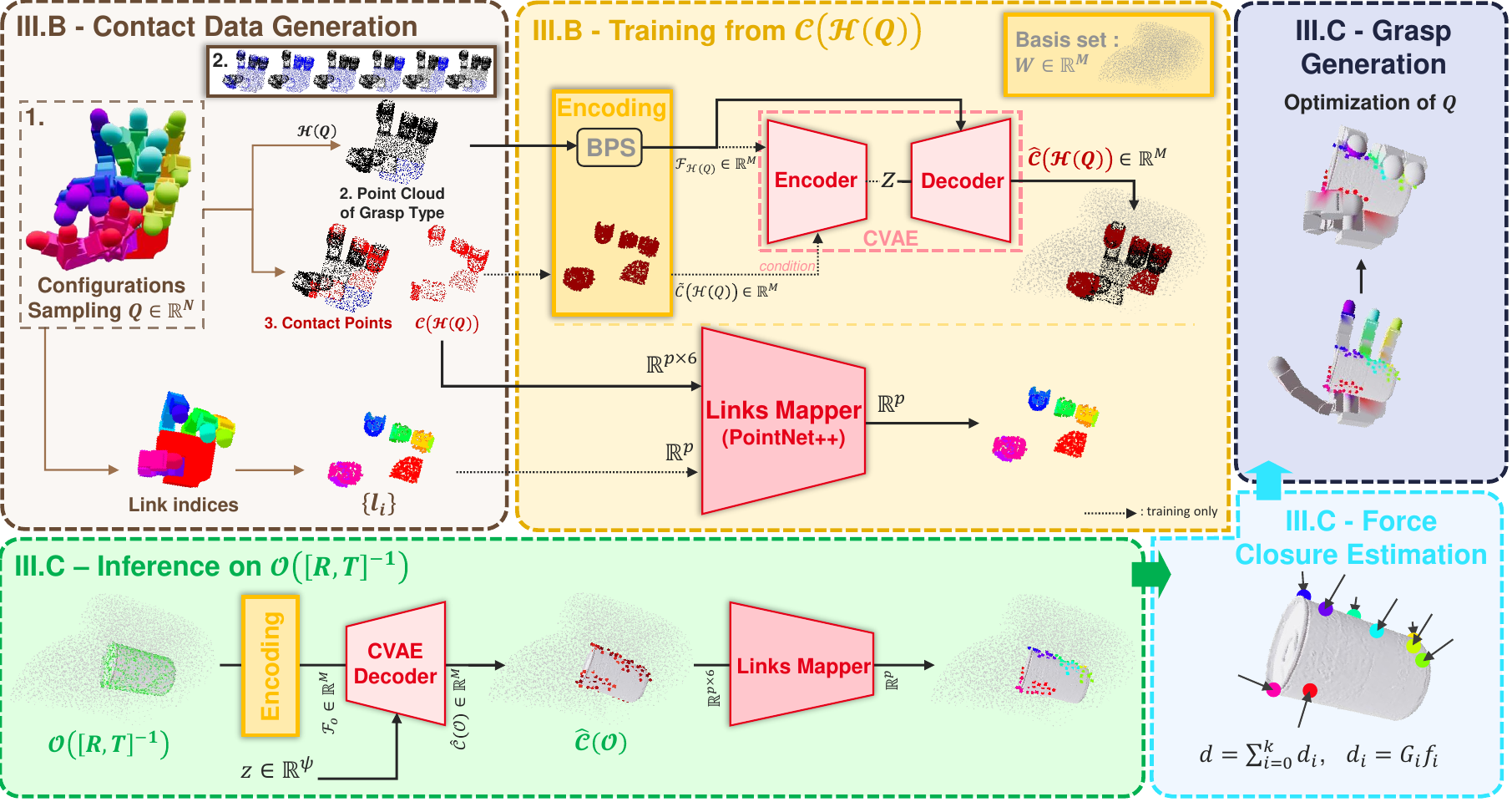}
    \caption{\textbf{Overview of \acronyme.}
    Geometrical graspability is learned in an object-agnostic manner by focusing on the gripper's capabilities. \textbf{Training:} We sample gripper configurations $Q$ to generate $\calH(Q)$ and corresponding contact points ($\calC(\calH(Q))$). To ensure transferability, we use a Basis Point Set (BPS) encoding tied to the gripper's workspace. A Conditional Variational Autoencoder (CVAE) is trained to reconstruct these contact distributions, while a Links Mapper (PointNet++) learns to associate contact points with specific gripper links. \textbf{Inference:} A novel object $\calO$, positioned at the inverse gripper pose $[R,T]^{-1}$, is BPS-encoded. By sampling a latent variable $z \in \mathbb{R}^\psi$, the CVAE Decoder generatively predicts diverse, plausible contact points $\widehat{\calC}(\calO)$. The Links Mapper then labels which gripper link should reach each point. \textbf{Generation:} Finally, a Force Closure check ensures the predicted contacts yield a stable grasp, and a Grasp Optimization step outputs the final, refined gripper configuration $Q^*$.
    }
    \label{fig:method_overview}
    \vspace{-10pt}
\end{figure*}

Our goal is to estimate numerous grasps when presented with a specific object shape.
The specificity of our approach is that the training phase only considers the gripper geometry, while the inference phase only considers the targeted object geometry. The underlying notion is a shift in perspective regarding the contact region: instead of defining it on the object, we define it as an intrinsic property of the gripper.

For the remainder of this section, we fix the pose of the gripper, defined by the rotation matrix $R \in $ SO(3), and the translation vector $T \in \mathbb{R}^3$, relative to the reference frame of the object $\calO$. We characterize a grasp by the pose of the gripper $[R,T]$ coupled with its joint values, $Q$. We define the gripper handprint $\calH = \{h_i\}$ as the set of points on the gripper's surface that represent its intrinsic contact capabilities, located on the active grasping surfaces (i.e., the palm and the inward-facing surfaces of the links).

\subsection{Shifting to a Gripper-Oriented Paradigm}
\label{subsec:gdoag_paradigm}

Given a gripper handprint $\calH = \{h_i\}$ and an object point cloud $\calO = \{o_i\}$, the set of contact points $\calC(.)$ is typically defined within the object's coordinate frame and consists of points on the object's surface that fulfill a certain criterion of distance with the gripper. Following~\cite{li2023gendexgrasp}, we express it as:
\begin{multline}
\calC(\calO) = \\
\{o_i \in \calO, \exists h_j \in \calH(R,T,Q) ~|~ \calD_{aligned}(o_i, h_j) < \epsilon\},
\label{eq:objectContactPointDefinition}
\end{multline}
where $\calH(R,T,Q)$ represents the gripper handprint positioned at the grasp pose $[R,T]$ with the joint configuration $Q$, and where the function $\calD_{aligned}$ is the aligned distance between any two points $x$ and $y$, as introduced by~\cite{li2023gendexgrasp}:
\begin{align}
    \calD_{aligned}(x, y) = e^{\gamma(1-\langle x-y, n_x\rangle)} \sqrt{\Vert x - y \Vert_2},
    \label{eq:aligned_distance}
\end{align}
with $n_x$ the surface normal at $x$ and $\gamma$ a scaling factor. \\
In our approach, we propose to recast this paradigm in an alternative gripper-oriented viewpoint: By applying the inverse transformation $[R,T]^{-1}$ to the object, we can analyze the grasp in the gripper's canonical coordinate frame. This refraiming allows us to define the contact points as a subset of the gripper's surface rather than the object's:
\begin{multline}
\calC(\calH(Q)) = \\
\{h_i \in \calH(Q), \exists o_j \in \calO([R,T]^{-1}) ~|~ \calD_{aligned}(h_i, o_j) < \epsilon\},
\label{eq:gripperContactPointDefinition}
\end{multline}
where $\calH(Q)$ is the gripper in its default pose (at origin with no rotation) and joint configuration $Q$, while $\calO([R,T]^{-1})$ is the object transformed into the gripper's frame. \\
While Equation~\ref{eq:gripperContactPointDefinition} defines the physical condition for contact with a specific object, the set of all kinematically feasible contact patterns is bounded by the gripper's geometry and constraints. \\
\textbf{Object-Agnostic Contact Priors.} Our key insight is that we can learn this intrinsic distribution of feasible contact zones solely from the gripper's kinematics and grasp taxonomy. This shifts the problem from finding specific contacts for a specific object (which requires $\calO$) to learning the gripper's intrinsic contact priors (which is object-independent). Consequently, we can pre-compute a manifold of valid contact surfaces directly on the gripper in a multitude of configurations, which are then queried against object shapes only at inference time.

\subsection{Training \acronyme from $\calC(\calH(Q))$}
\label{subsec:HOIlearning}

We build on this observation to formulate our object-agnostic training procedure. The training process requires generating diverse gripper configurations, encoding them into a consistent spatial representation, and training two parallel networks to predict contact distributions and kinematic assignments.
The following architecture has been explored further in~\cite{merand2026cotograsp}.

\noindent\textbf{Contact Data Generation.}
The training data is generated exclusively from the gripper’s kinematics and an adapted taxonomy, as illustrated in Figure~\ref{fig:grasp_types}. We first uniformly sample valid joint configurations $Q$ within the gripper's kinematic limits. For each configuration, we randomly select a grasp type from the taxonomy presented in~\cite{escorcia2023task}, which defines specific admissible contact regions on the gripper surface. Finally, we generate the target contact map $\calC(\calH(Q))$ by randomly sampling active contact points strictly within these admissible regions. While unconstrained random sampling could theoretically be used to populate the dataset, leveraging a taxonomy ensures that every generated sample is structurally valid. It acts as a crucial prior to guarantee that the sampled contact points exhibit a correct kinematic harmony, representing realistic, synergistic grasps rather than arbitrary and independent surface contacts. This procedural generation creates a diverse dataset of kinematically feasible contact distributions independent of any object geometry.

\noindent\textbf{BPS Encoding and Distance Field.}
To allow our neural networks to process variable-length point clouds efficiently, we project our input data -- denoted generally as $P$, which represents the gripper point cloud $\calH(Q)$ during training or the object point cloud $\calO$ during inference -- into a fixed-size representation using a specialized Basis Point Set~\cite{prokudin2019efficient} (BPS). 
Instead of a standard bounding box, we discretize the gripper’s kinematic workspace to form our BPS, denoted as $W \in \mathbb{R}^{M \times 3}$. This workspace represents the volume of points reachable by the gripper relative to a fixed palm frame, naturally concentrating the data where grasps occur.
This projection implies an inherent train-test domain shift between the training ($\calH(Q)$) and testing ($\calO$) geometry, a cross-domain transfer challenge that has been explored further in \cite{merand2026cotograsp}.
For any input point cloud $P \in \mathbb{R}^{p \times 6}$, comprising 3D spatial coordinates concatenated with their corresponding 3D surface normals, we compute an aligned distance field $\calF_{P} \in \mathbb{R}^{M}$. Following \cite{li2023gendexgrasp}, the distance from each basis point $w_i \in W$ to the point cloud $P$ is defined as:
\begin{equation}
    \calD(P, w_i) = \min_{p_j\in P} \calD_{aligned}(p_j, w_i).
    \label{eq:bps_distance}
\end{equation}
To provide a supervision signal, we define the projected contact field $\tilde{\calC}(P) = \{c_i \}_{i=1}^{M}$. Each component $c_i \in [0, 1]$ quantifies the proximity and alignment of each $w_i$ to the active contact points $\calC(P)$. Following \cite{jiang2021hand}, this is computed as:
\begin{equation}
c_i = 1 - 2 \times (\text{Sigmoid}(\calD(\calC(P), w_i)) - 0.5).
\end{equation}
This function provides a continuous "contact likelihood," approaching 1 when $w_i$ is close and aligned to a true contact point, and 0 when it is distant or misaligned.

\noindent\textbf{CVAE Architecture and Training.}
With our data structured into the BPS representation $\calF_{P}$ and contact labels $\tilde{\calC}(P)$, we train a Conditional Variational Autoencoder~\cite{sohn2015learning} (CVAE) to learn the distribution of feasible contacts.
The encoder takes the concatenated input $[\calF_{P}, \tilde{\calC}(P)]$ of shape $M \times 2$ and predicts the parameters of the posterior distribution $q_{\theta}(z \mid \calF_{P}, \tilde{\calC}(P))$. A latent variable $z \in \mathbb{R}^{\psi}$ is then sampled from this distribution. The decoder, following \cite{mayer2022ffhnet}, consists of two Fully Connected ResBlocks that take the concatenation of $z$ and $\calF_{P}$ to output predicted contact values $\widehat{\calC}(P)$. The network is optimized using the following loss function:
\begin{equation}
    L = L_{\text{recon}} + \beta D_{\text{KL}}(q_{\theta}(z \mid \calF_{P}, \tilde{\calC}(P)) ~\|~ \calN(0,I)).
\end{equation}
where $\beta$ weights the Kullback-Leibler divergence.
The reconstruction loss $L_{\text{recon}}$ is an $L_2$-Norm scaled by an attention weight $e^{\alpha c_i}$ to heavily prioritize the accuracy of high-likelihood contact regions:
\begin{equation}
    L_{\text{recon}} = \sqrt{\frac{\sum_{i=1}^M (c_i - \hat{c}_i)^{2} \times e^{\alpha c_i}}{\sum_{k=1}^M e^{\alpha c_i}}}.
\end{equation}

\noindent\textbf{Links Mapper.}
In parallel to the CVAE, a PointNet++~\cite{qi2017pointnet++} is trained on the sampled point clouds to associate each point $p_i \in \calC(P)$ with the corresponding gripper phalanx link $l_i$. This assigns kinematic meaning to the spatial points, which is crucial for retrieving the full gripper joint configuration during the downstream optimization phase.

\subsection{\acronyme Inference on $\calO([R,T]^{-1})$}
\label{subsec:HOIinference}

Once trained, the models can be deployed to predict contact zones on unseen objects. Given a novel object point cloud $\calO$ transformed into the gripper's canonical workspace via $[R,T]^{-1}$, we first compute its aligned distance field $\calF_{\calO}$ against the basis set $W$, identical to the training phase. \\
Bypassing the encoder, we sample a latent vector $z \sim \calN(0,I)$. The decoder takes the concatenation $[z, \calF_{\calO}]$ and predicts the continuous contact field $\widehat{\calC}(\calO)$. Instead of projecting these values back onto the object's surface, we operate directly within the workspace domain. We extract the intended contact locations by isolating the basis points that exhibit a high predicted contact likelihood ($\hat{c}_i > \tau$, where $\tau$ is a confidence threshold). This yields a discrete set of contact points $\calC(W)$. \\
Finally, the pre-trained Links Mapper evaluates $\calC(W)$ to assign each point to a specific gripper phalanx.
While this generates a geometrically plausible contact map, these predictions are sampled from a latent prior and do not guarantee physical stability. Consequently, the pipeline concludes with a rapid force-closure estimation followed by a full kinematic optimization to refine the final grasp execution, using $\calC(W)$ as the targets for the gripper's links.

\noindent\textbf{Contact Points Force-Closure Estimation.}
Unlike methods trained on pre-validated grasp datasets~\cite{li2023gendexgrasp, wei2024d}, our generative approach requires an explicit assessment of grasp stability. To ensure the inferred contact points can theoretically yield a stable grasp before performing the computationally expensive joint configuration optimization, we evaluate the Force-Closure condition directly on the predicted contact points $\calC(W)$.

\noindent a) \underline{\textit{Contact Modeling:}} We first group the predicted contact points by their associated gripper phalanx indices. To enforce a unique contact location per phalanx, we compute the barycenter of each cluster and project it onto the object’s surface, yielding a set of discrete contact locations $b_i$. We assume a Coulomb friction model with a friction coefficient of $\mu = 0.3$. As this estimation is performed in the gripper's reference frame to assess geometric feasibility, we do not consider the gravity or object's weight.

\noindent b) \underline{\textit{Force-Closure Computation:}} We compute the grasp wrench space, defined as the convex hull of all possible wrenches generated by unit contact forces at locations $b_i$. The total wrench is given by:
\begin{equation} 
\label{eq:wrench}
    d = \sum_{i=1}^{k} d_i = \sum_{i=1}^{k} G_i f_i
\end{equation}
where $G_i$ is the partial grasp matrix for contact $b_i$, and $f_i$ represents the primitive force vectors along the edges of the friction cone, normalized such that $\|f_i\| = 1$. We verify the force-closure condition by checking if the origin of the wrench space lies strictly within the interior of this convex hull.

\noindent c) \underline{\textit{Resampling Strategy:}} If the force-closure condition is not met, it implies the predicted contact distribution is unstable. In this case, we discard the prediction and sample a new latent variable $z \in \mathbb{R}^{\psi}$ to generate a fresh $\hat{\calC}(\calO)$. To prevent exhaustive searches when an object pose ($[R,T]^{-1}$) is poorly conditioned, specifically when it falls near the boundaries of the gripper's workspace where valid contacts are scarce, we limit this resampling process to a maximum of 20 iterations.
It is important to note that this step validates the stability potential of the contact configuration using the simplified barycenter model. The final physical grasp quality is determined by the gripper’s ability to reach these point during the subsequent optimization, which optimizes the gripper to fit the full dense contact regions rather than just these discrete barycenters.

\noindent \textbf{Grasp Generation.}
Determining the gripper joint configuration $Q^*$ that enables contact with the inferred object contact points is formulated as an optimization problem under non-penetration constraints. While some methods minimize the Euclidean distance between a predicted grasp and the gripper's actual link transformations \cite{wei2024d} or between specific points on the gripper and the object \cite{khargonkar2025robotfingerprint}, this often lacks the necessary mechanisms to prevent hand-object and self-penetrations, which are essential for physical feasibility. \\
Instead, building on approaches like \cite{khargonkar2025robotfingerprint, li2023gendexgrasp}, we determine the optimal joint configuration $Q^*$ by aligning the gripper with the inferred contact points while strictly penalizing collisions. We achieve this by minimizing the following energy function:
\begin{equation}
\label{eq:energy_opti}
   Q^* = \operatorname*{argmin}_{Q} \left( \lambda_d E_{dist} + \lambda_p E_{pen} + \lambda_s E_{spen} + \lambda_j E_{j} \right)
\end{equation}
where $\lambda$ terms are weighting coefficients set to $\lambda_d = 1.0$, $\lambda_p = 50.0$, $\lambda_s = 0.1$, $\lambda_j = 1.0$. The components are defined as follows:

\noindent a) \underline{\textit{Contact Distance ($E_{dist}$):}}. We minimize the distance from the predicted contact points $\calC(W)$ on the object to their assigned gripper links. Utilizing the link indices predicted by the Links Mapper (Section~\ref{subsec:HOIlearning}), we define:
$$E_{dist} = \sum_{p \in \calC(W)} \min_{h \in \calH_{l(p)}(Q)} \| p - h \|_2^2,$$
where $\calH_{l(p)}(Q)$ represents the subset of handprint points belonging to the specific link $l(p)$ assigned to contact point $p \in \calC(W)$. This ensures the gripper fingers move precisely to the target regions.

\noindent b) \underline{\textit{Object Penetration ($E_{pen}$):}} To prevent the gripper from intersecting the object volume, we penalize gripper points that penetrate the object's signed distance field, denoted as $\Phi_{\calO}(\cdot)$:
$$E_{pen} = \sum_{h \in \calH(Q)} \max(0, - \Phi_{\calO}(h))$$

\noindent c) \underline{\textit{Self-Penetration ($E_{spen}$):}} We enforce self-collision avoidance by penalizing distances between disjoint gripper links $i$ and $j$ that fall below a safety threshold $\epsilon$ (set to $0.025$m):
$$E_{spen} = \sum_{i,j} \max(0, \epsilon - \operatorname{dist}(k_i(Q), k_j(Q)))^2,$$
where $k_i(Q)$ denotes the centroid of the bounding box for link $i$ in configuration $Q$.

\noindent d) \underline{\textit{Joint Limits ($E_j$):}} Finally, we constrain the joints to remain within their hardware kinematic limits $[Q_{min}, Q_{max}]$:
$$E_{j} = \| \max(0, Q - Q_{max}) \| + \| \max(0, Q_{min} - Q) \|$$
\section{Experiments}

\begin{table*}[thbp]
    \centering
    \renewcommand\arraystretch{1.25}
    \resizebox{\textwidth}{!}{
        \begin{threeparttable}
            \begin{tabular}{l|cc|ccc|c|ccc|ccc}
                \toprule
                \multirow{2}{*}{\textbf{Method}}
                & \multirow{2}{*}{\begin{tabular}[c]{@{}c@{}}\textbf{Data} \\ \textbf{Driven} \end{tabular}}
                & \multirow{2}{*}{\begin{tabular}[c]{@{}c@{}}\textbf{Object-Agnostic} \\ \textbf{Training} \end{tabular}}
                & \multicolumn{4}{c|}{\textbf{Success Rate (\%) $\uparrow$}}
                & \multicolumn{3}{c|}{\textbf{Efficiency (sec. / grasps) $\downarrow$}}
                & \multicolumn{3}{c}{\textbf{Diversity (avg.) $\uparrow$}}
                \\ 
                \cline{4-13} 
                & & & Barrett & Allegro & ShadowHand & Avg.
                 & Barrett & Allegro & ShadowHand
                 & T (m) & R (rad) & Q (rad)
                \\ \cline{1-13} 
                DFC~\cite{liu2021synthesizing}
                    & \xmark & \cmark
                    & 83.10 & 82.71 & 72.15 & 79.32
                    & >1800 & >1800 & >1800
                    & \textbf{0.0607} & 1.424 & \textbf{0.3579}
                \\
                GenDexGrasp~\cite{li2023gendexgrasp} (full)
                    & \cmark & \xmark
                    & 70.26 & 71.48 & 71.15 & 70.96
                    & 9.78 & 16.45 & 14.65
                    & 0.0519 & 1.416 & 0.2567
                \\
                DRO-Grasp~\cite{wei2024d} (pretrain, w/o controller)
                    & \cmark & \xmark
                    & 78.30 & 75.80 & 63.30 & 72.47
                    & 0.88 & 0.42 & 1.72
                    & 0.0546 & \textbf{1.515} & 0.2892
                \\ \cline{1-13}
                \textbf{\acronyme} (w/o FC)
                    & \cmark & \cmark
                    & 86.30 & 91.20 & 74.70 & 84.07
                    & \textbf{0.09} & \textbf{0.13} & \textbf{0.15}
                    & 0.0480 & 1.396 & 0.3162
                \\
                \textbf{\acronyme}
                    & \cmark & \cmark
                    & \textbf{87.40} & \textbf{93.20} & \textbf{77.90} & \textbf{86.93}
                    & 0.18 & 0.19 & 0.20
                    & 0.0479 & 1.401 & 0.3170
                \\
                \bottomrule
            \end{tabular}
        \end{threeparttable}
    }
    \caption{\textbf{In-depth grasp performance analysis on the Multidex~\cite{li2023gendexgrasp} test set.} We report our grasp results, on three classic dexterous grippers in terms of success rate, efficiency and diversity.}
    \label{tab:result}
    \vspace{-15pt}
\end{table*}

\subsection{Implementation Details}
\label{subsec:implementation}

\noindent\textbf{Training Data Generation.}
To ensure dense coverage of the gripper's reachable workspace, we uniformly sampled $10,000$ kinematically valid gripper configurations $Q$ and computed their corresponding surface point clouds $\calH(Q)$. 
These active point clouds are extracted from the full mesh by applying a gripper-specific threshold to the dot product of each vertex normal and the palm's facing direction, effectively isolating the primary contact pads and the immediate lateral sides of the fingers.
We utilized the grasp taxonomy adapted from~\cite{escorcia2023task}, selecting the 6 most common and suitable grasp types for robotic grippers, as detailed in Figure~\ref{fig:grasp_types}. According to the literature \cite{zheng2011investigation, gonzalez2014analysis}, these grasp types are utilized for $92.5\%$ of the total working time by machinists and $96.2\%$ of the time by housemaids.
For each configuration and its assigned grasp type, we uniformly sampled 50 sets of admissible contact points strictly within each active regions defined by the taxonomy. This process yielded a dataset of $3,000,000$ labeled point clouds. 

\noindent\textbf{Computational Efficiency.}
A significant advantage of our object-agnostic formulation is the speed of data generation. The entire dataset creation required approximately 1 GPU hour on a single Nvidia RTX 4090. In contrast, a previous work~\cite{li2023gendexgrasp} reported a much longer generation time of 1,400 GPU hours using Nvidia A100.

\noindent\textbf{Network Hyperparameters.}
We discretized the workspace $W$ using $M=8,192$ points, set the scaling factor $\gamma=2.0$ for the aligned distance (Eq.~\ref{eq:aligned_distance}) and used $\tau = 0.8$ as confidence threshold. The CVAE was trained with a latent dimension size of $\psi=128$, using a weighting parameter $\beta=0.01$ and an attention factor $\alpha=3.0$. We trained the CVAE for 100 epochs and the PointNet++ links mapper for 50 epochs. Training was parallelized across multiple Nvidia A100 GPUs with a batch size of 128.

\subsection{Grasp Pose Constraints and Generation}
\label{subsec:pose_generation}

A key advantage of this method is the flexibility to impose the object's pose $[R,T]$. In a practical grasping scenario, the transformation between the gripper and the object is typically constrained by the task and the physical environment. Consequently, generating grasp poses that are not physically feasible is inefficient. Furthermore, objects with translational or rotational symmetries (e.g., a cylinder rotating around its axis of revolution) can be grasped effectively with a single pose inference, as it provides a broad range of valid grasp configurations for that object type.

\noindent\textbf{Grasp Pose Generation Strategy}. 
To address the need for comprehensive grasp coverage, particularly for objects lacking significant symmetry, we propose a strategy, founded on \cite{wang2023dexgraspnet}, for generating grasp candidates. We sample gripper poses uniformly on the object's convex hull, which is dilated by 110\%. The gripper's palm is oriented to point opposite to the hull's normal vector. This approach ensures that the generated grasp poses are both diverse and well-distributed across the object's surface.

\subsection{Evaluation Metrics}
\label{subsec:metrics}

\noindent\textbf{Success Rate:} We quantify grasp success using the Isaac Gym~\cite{makoviychuk2021isaac} simulator. We initialize the object and gripper in the optimized configuration $Q^*$ and, similar to \cite{li2023gendexgrasp}, apply sequential external forces on the object along the $\pm \text{xyz}$ directions for one second each. A grasp is considered successful if the object deviates less than 2 cm from its original pose. \\
This simulation-based verification is essential because the prior Force Closure Estimation (Section~\ref{subsec:HOIinference}) validates only the theoretical stability of simplified barycenters. The subsequent Joint Optimization (Section~\ref{subsec:HOIinference}) is the first step to incorporate the full object geometry and gripper kinematics. Consequently, contact distributions deemed valid by the force-closure estimation may prove kinematically unreachable or geometrically obstructed by the object's shape in practice, preventing a perfect match and necessitating physical validation of the final grasp.

\noindent\textbf{Efficiency:} We evaluate the efficiency of all models by calculating the average time required to generate a single grasp. This measurement is based on the total time to produce 100 grasps, including both the model inference and optimization phases. The resulting value is then normalized to a single grasp. Following \cite{wei2024d}, the time for the Isaac Gym simulation is excluded from this metric.

\noindent\textbf{Diversity:}  
We measure diversity as the standard deviation of the gripper position ($T$), orientation ($R$), and joint values ($Q$). Although the candidate gripper poses ($R, T$) are initially sampled, the variability observed in successful grasps is not arbitrary; it is intrinsically shaped by the object's specific geometry and the gripper’s kinematic capabilities, which naturally filter the reachable contact manifold. We therefore report diversity across all components to demonstrate this geometric adaptation, while specifically highlighting the variance in $Q$ to verify our model’s generative ability to synthesize distinct finger configurations for similar poses.

\subsection{Experimental Protocols}
\label{subsec:exp_protocols}

\begin{figure}[!t]
    \centering
    \includegraphics[width=0.9\linewidth]{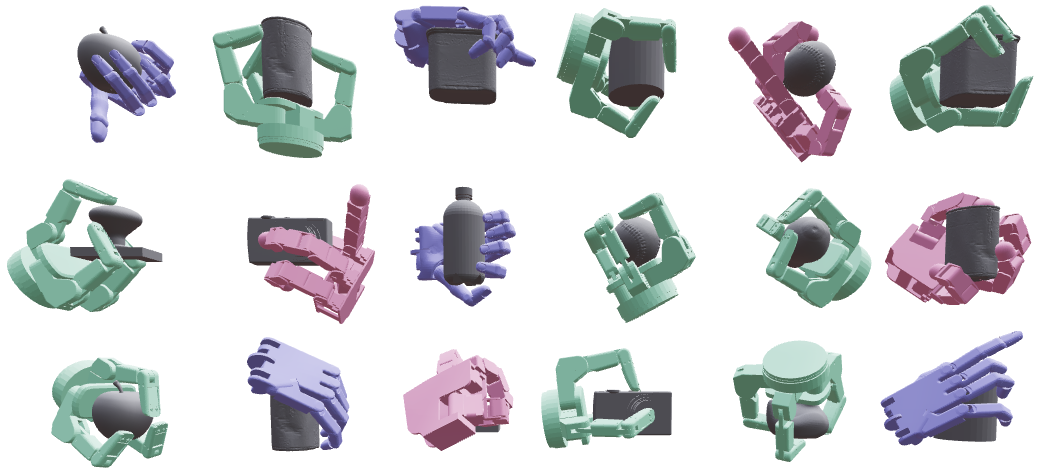}
    \caption{\textbf{\acronyme grasp results on Multidex~\cite{li2023gendexgrasp} objects.
    } Grasps are shown for the Barrett (green), Allegro (pink), and Shadow Hand (purple) grippers.}
    \label{fig:exp_simu}
    \vspace{-13pt}
\end{figure}

\begin{table*}[!t]
    \centering
    \begin{tabular}{lc|ccccc|c}
        \toprule
        \textbf{Method} & \textbf{Per-Dataset Training} & \textbf{DexGraspNet $\uparrow$} & \textbf{UniDexGrasp $\uparrow$} & \textbf{MultiDex $\uparrow$} & \textbf{RealDex $\uparrow$} & \textbf{DexGRAB $\uparrow$} & \textbf{Avg. (\%)} \\
        \midrule
        UniDexGrasp~\cite{xu2023unidexgrasp} & \cmark & 33.9 & 23.7 & 21.6 & 27.1 & 20.8 & 25.42\\
        GraspTTA~\cite{jiang2021hand} & \cmark & 18.6 & 21.0 & 30.3 & 13.3 & 14.4 & 19.52 \\
        SceneDiffuser~\cite{huang2023diffusion}& \cmark  & 26.6 & 28.3 & 69.8 & 21.7 & 39.1 & 37.10 \\
        UGG~\cite{lu2024ugg} & \cmark & \underline{46.9} & 46.0 & 55.3 & 32.7 & 42.7 & 44.72 \\
        DGA~\cite{zhong2025dexgrasp} & \cmark & \textbf{57.5} & \textbf{53.1} & \textbf{79.1} & \textbf{44.8} & 57.9 & \textbf{58.48} \\
        \midrule
        \textbf{\acronyme} & \xmark & 43.07 & \underline{49.51} & \underline{77.90} & \underline{37.37} & \textbf{62.13} & \underline{53.97} \\
        \bottomrule
    \end{tabular}
    \caption{\textbf{Grasp generalization assessment across multiple grasp datasets.} Following \cite{zhong2025dexgrasp} we evaluate our method performances with the Shadow hand on multiple grasp test sets. We report the success rates as defined in~\ref{subsec:metrics}. It is worth noting that state-of-the-art methods are retrained for each dataset. \acronyme has only been trained once on Shadow hand kinematics.}
    \label{tab:comparison}
    \vspace{-10pt}
\end{table*}

\textbf{In-depth grasp performance analysis.} We evaluated our method and several baselines on three distinct robotic grippers: the Barrett Hand~\cite{townsend2000barretthand}, Allegro Hand~\cite{allegro}, and Shadow Hand~\cite{tuffield2003shadow}. Our evaluation involved 100 inference runs per method, using a batch size of 10 to manage GPU memory.
We extensively compare \acronyme against the following baselines using the test set of the Multidex~\cite{li2023gendexgrasp} dataset. It comprises 10 objects from the ContactDB~\cite{brahmbhatt2019contactdb} and YCB~\cite{calli2015ycb} object sets.
We evaluated DFC~\cite{liu2021synthesizing} on grasps from the CMapDataset. The grasp poses were originally generated using DFC's method and subsequently refined, in~\cite{li2023gendexgrasp}, through a post-filtering process to get a clean dataset. This led to a higher success rate than what was originally reported.
We used a pre-trained model checkpoint of GenDexGrasp~\cite{li2023gendexgrasp} with data from multiple robot hands and default hyperparameters. We first generated contact maps for the object set and then derived the grasp poses using the optimization framework provided by the authors.
We selected the most effective version of DRO-Grasp~\cite{wei2024d}, which includes configuration-invariant pre-training. We kept all hyperparameters at their default values and deactivated the simple grasp controller for a fair comparison.

\textbf{Generalization across multiple grasp datasets.} To extend the evaluation of our method on a wide range of objects, we followed \cite{zhong2025dexgrasp} by using the test set of three simulated datasets~\cite{li2023gendexgrasp, wang2023dexgraspnet, xu2023unidexgrasp}, a real-world dataset~\cite{liu2024realdex} and a human hand dataset~\cite{taheri2020grab} retargeted to dexterous hand parameters by~\cite{zhong2025dexgrasp}. This makes a total of $3438$ objects from five different benchmarks.
The evaluation protocol remains consistent with the procedure described above, with all experiments conducted exclusively using the Shadow Hand.

\subsection{Overall Performances}
\label{subsec:overall_performances}

\textbf{In-depth grasp performance analysis.} On the Multidex dataset (Table~\ref{tab:result}), our method, \acronyme, demonstrates superior performance. While being trained in an object-agnostic manner, \acronyme achieved a higher average success rate for generating accurate grasps across all three grippers compared to the baselines. 
This high accuracy is paired with good efficiency. While DRO-Grasp~\cite{wei2024d} may appear faster for a single grasp, its processing time scales linearly with the number of grasps, as each is optimized independently. Our method, in contrast, benefits from a vectorized optimization process that minimizes a single energy function for all grasp candidates simultaneously. This approach, while having a higher initial overhead, results in a more consistent and significantly faster processing time when generating a large number of grasps. Furthermore, our network is more compact than DRO-Grasp's in terms of parameter count. Compared to GenDexGrasp~\cite{li2023gendexgrasp}, which does not vectorize its optimization, our approach gains a substantial efficiency advantage.
The strong performance of our model even without an explicit force closure estimation confirms a key design principle: the model successfully learns to map gripper capabilities and shapes onto objects. This demonstrates that the expressiveness of our training set enables the model to implicitly learn robust grasp mechanics.
Because our model is gripper-centric, the diversity of grasps for a given object depends on how comprehensively we sample object poses, rather than being a property of the network's output itself. However, our model can still generate multiple grasp types for a single object pose by sampling from the CVAE's latent space, providing a means for diverse grasp synthesis.

\textbf{Generalization across multiple grasp datasets.} We report the success rates achieved across all five benchmarks using the Shadow Hand~\cite{tuffield2003shadow} in Table~\ref{tab:comparison}. Our method achieved the second-highest average success rate across all five datasets. This is particularly noteworthy because all competing methods were specifically trained on each respective dataset, whereas \acronyme was trained only once on the Shadow Hand in an object-agnostic manner. These results were achieved using a simple pose sampling method (\ref{subsec:pose_generation}), which is well-suited for objects with a volume enclosed within the gripper's workspace. For larger objects, however, this approach is less effective. Adopting a more advanced sampling strategy specifically adapted for such objects would likely yield stronger grasp performance.

\begin{figure}[h]
    \centering
    \resizebox{0.9\linewidth}{!}{
        \hbox{
            \includegraphics[width=0.55\linewidth]{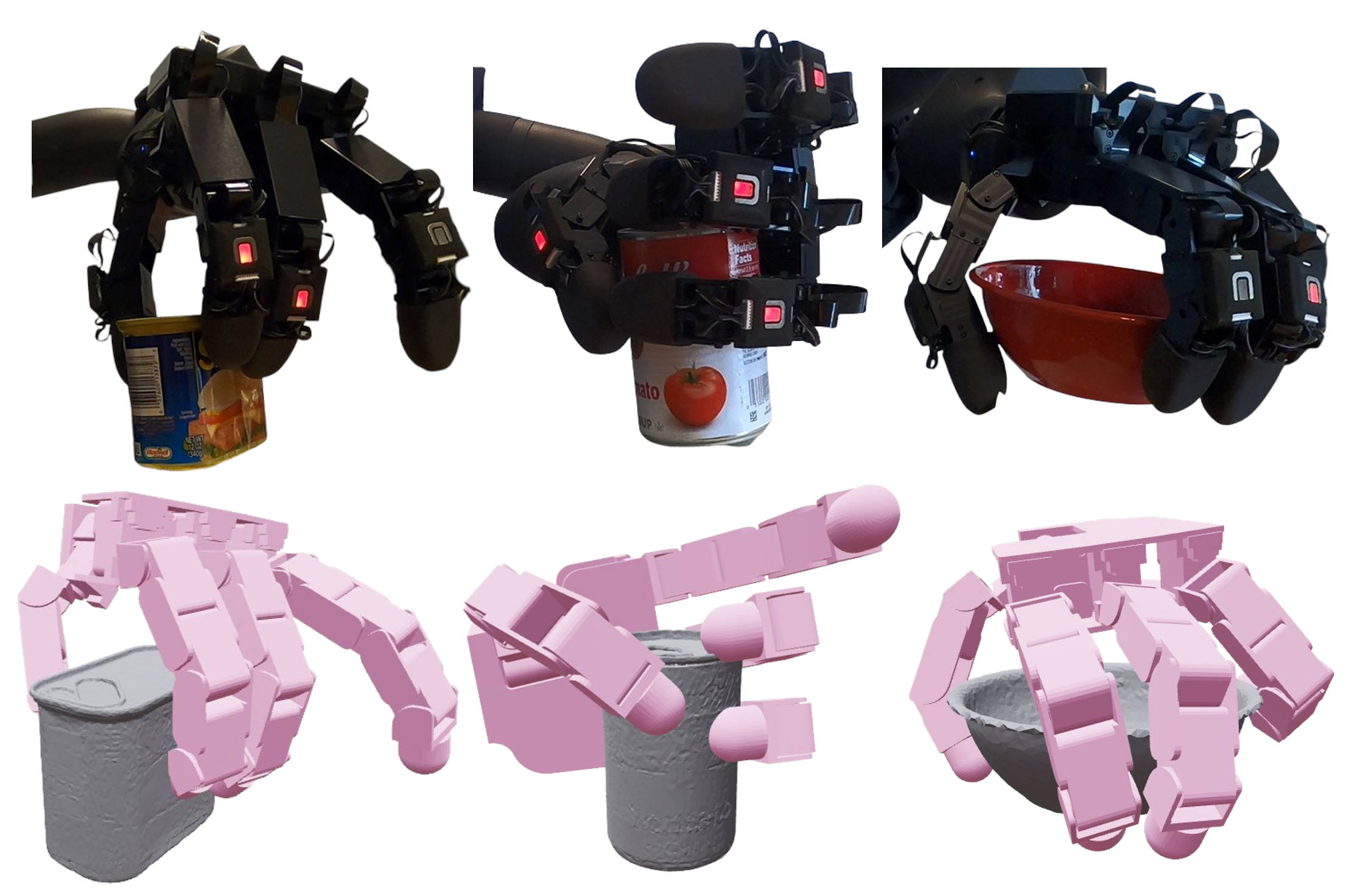}
            \hfill
            \includegraphics[width=0.4\linewidth]{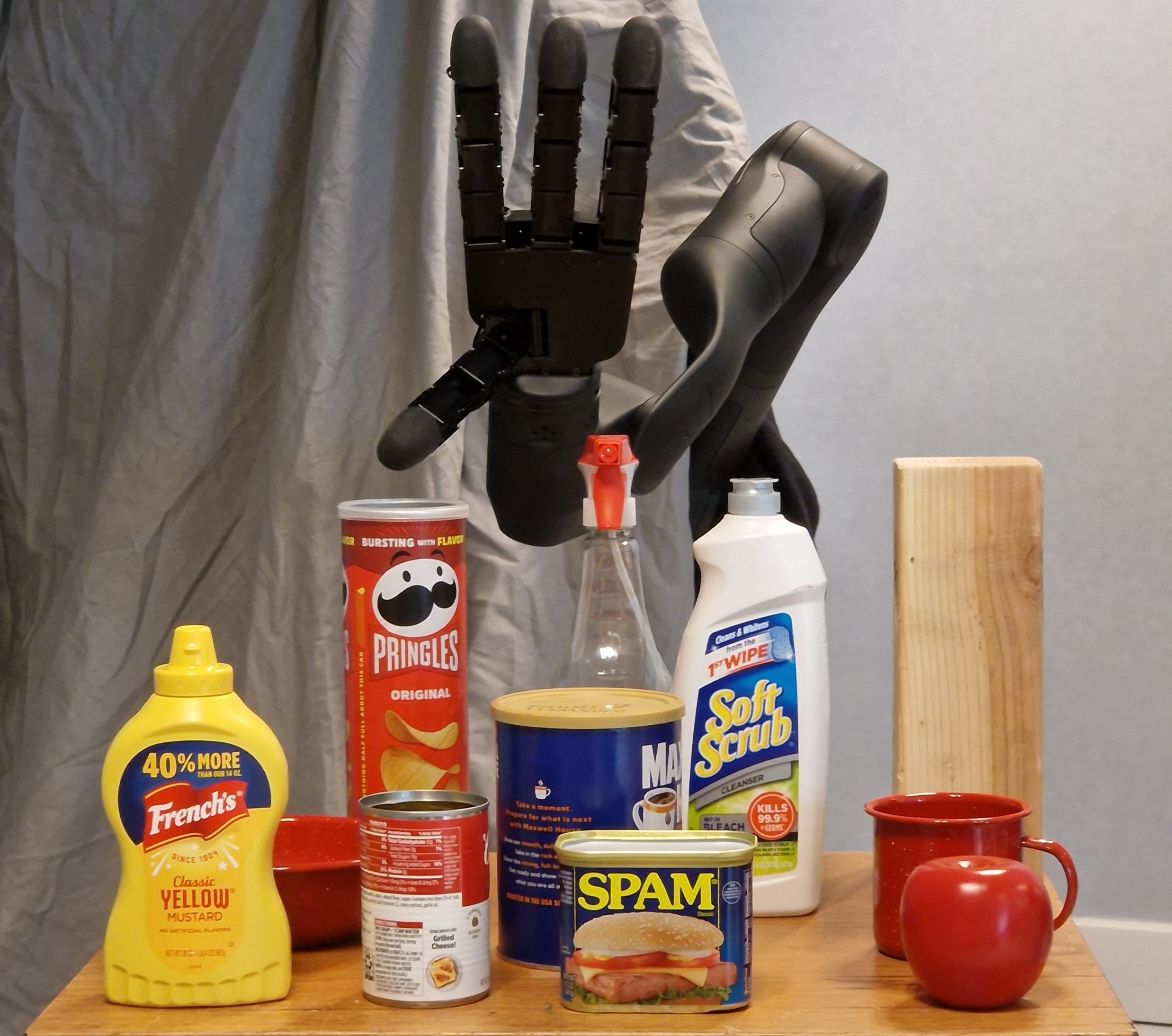}
        }
    }
    \caption{\textbf{Real-world setup and results with Allegro hand on YCB~\cite{calli2015ycb} objects.}
    First row presents the real robot grasps. Second row presents corresponding virtual grasps. Objects have been rotated around the z-axis for a better understanding of the grasp poses.
    }
    \label{fig:exp_realvssimu_setup}
    \vspace{-10pt}
\end{figure}

\subsection{Real-Robot Experiments}

We conducted experiments using an Allegro Left Hand mounted on a 7-DoF robot arm.
We successfully grasped 11 objects from the YCB Dataset, demonstrating the method's ability to transfer to real-world objects. Successful grasps are illustrated Figure~\ref{fig:exp_realvssimu_setup} and the setup is shown Figure~\ref{fig:exp_realvssimu_setup}.
Videos of the experiments are provided in supplementary material.
\section{Conclusion}

This paper introduces \acronyme, a novel learning paradigm for data-driven grasp planners. Our approach leverages the strengths of deep learning while eliminating the need for a large-scale, object-specific grasp database. By adopting a gripper-centric training phase that makes no assumptions about object shapes, our model learns a generalizable grasp strategy. Extensive evaluations across multiple benchmarks demonstrate that our method achieves performance competitive with state-of-the-art approaches, even without training on their specific datasets. This highlights the strong generalization capabilities of our formulation, which we also validate with a successful real-robot deployment.


\bibliographystyle{IEEEtran}
\bibliography{refs}

\end{document}